\documentclass[11pt]{article}

\usepackage[preprint]{acl}

\usepackage{times}
\usepackage{latexsym}
\usepackage[T1]{fontenc}
\usepackage[utf8]{inputenc}
\usepackage{microtype}
\usepackage{inconsolata}
\usepackage{graphicx}
\usepackage{booktabs}
\usepackage{array}
\usepackage{amsmath}

\newcommand{\dnew}{\ensuremath{\Delta_{\mathrm{new}}}}
\newcommand{\dold}{\ensuremath{\Delta_{\mathrm{old}}}}

\title{Retrieved but Not Read: 
Understanding Why Search-Augmented LLMs Fail to Use Available Evidence
}

\author{Owen Walters\thanks{Work done while at Imperial College London.} \qquad Lihu Chen \qquad Francesca Toni \\
  Department of Computing, Imperial College London, UK \\
  \texttt{\{lihu.chen, ft\}@imperial.ac.uk}}
\begin{document}
\maketitle

\begin{abstract}
Web search allows LLMs to access up-to-date information,
but it remains unclear why they sometimes produce
outdated answers even when relevant evidence is available.
To study this problem, we construct temporal facts
whose values changed after the models' training cutoffs.
We evaluate one proprietary model under deployed web search and six open-weight model configurations under controlled retrieval.
We find that models often answer from search snippets
without opening the source pages, which is associated with
higher rates of outdated answers and unfaithful citations.
Our controlled experiments and layer-wise analysis
reveal a gap between the availability of evidence and its effective use.
Even when retrieved evidence influences internal
predictions, models may still favour outdated
information in their final answers.
We hope these findings will help advance research on the reliable use of retrieved evidence in search-augmented LLMs.

\end{abstract}

\section{Introduction}
Deployed LLM assistants such as ChatGPT~\citep{openai-2024-search} often use web search to access
up-to-date information.
However, retrieving relevant search results
does not guarantee that models use the information correctly.
Even when updated evidence is available, search-augmented LLMs may still produce
outdated answers or cite sources that contradict their responses (Figure~\ref{fig:teaser}).
This matters because citations lend answers provenance.
This raises an important question:
\emph{Why do search-augmented LLMs fail to use available evidence?}

\begin{figure}[t]
\centering
\small
\fbox{\parbox{\dimexpr\columnwidth-2\fboxsep-2\fboxrule\relax}{\raggedright
\textbf{Prompt.} \texttt{Today's date is 2026-08-08. Is the following accurate? \textquotesingle Alex Chriss is the CEO of PayPal.\textquotesingle} [answer-line instruction]\\[3pt]
\textbf{Trace.} One search (\texttt{site:paypal.com Alex Chriss CEO PayPal leadership}); 18 results returned; 0 pages opened.\\[3pt]
\textbf{Reply.} \texttt{Yes. \ldots{} Alex Chriss is PayPal's President and CEO. (paypal.com)\ \ FINAL: ACCURATE}\\[3pt]
\textbf{Ground truth.} Enrique Lores became PayPal's CEO on 1 March 2026.}}
\caption{
An example of the ``retrieved but not read'' phenomenon in a deployed search-augmented LLM.
The model searches once, opens no page, endorses a claim five months out of date and attributes it to an outdated article on the company's own site.}
\label{fig:teaser}
\end{figure}

Prior work has studied temporal knowledge~\citep{mousavi-etal-2024-dyknow, zhu-etal-2025-evolvebench}, citation faithfulness~\citep{wallat-etal-2025-correctness} and failures in retrieval-augmented generation (RAG)~\citep{chen-etal-2024-benchmarking, xie-etal-2024-adaptive}.
However, how search-augmented LLMs use web search results has received less attention, and it is unclear why these models fail even when relevant evidence is available.

To this end, we construct a dataset of \emph{post-cutoff temporal facts}, whose values changed after the deployed model's training cutoff.
We build a deterministic and regenerable pipeline that generates these facts from primary sources, and draw 225 facts from its output, across six domains including sports, finance and medicine.\footnote{Code and data: \url{https://github.com/ojwwalters/retrieved-but-not-read/}}
Each fact has an outdated value and a current value.
This lets us identify outdated answers, test whether models use updated evidence correctly and check whether the cited pages support the models' claims.

We first study a proprietary LLM under its native web search
to identify failures in a deployed setting.
We then test whether these failures recur under controlled
retrieval, giving the same proprietary LLM
and six open-weight model configurations an identical menu of frozen search results.
Our experiments reveal two main findings.
First, models frequently answer from result snippets without opening a page, and most stale answers and unfaithful citations occur in these episodes.
These failures are also shared across models (\S\ref{sec:results}).
Second, to understand why these failures happen, we run controlled experiments and a mechanistic analysis.
Providing the source page directly reduces some errors.
In the remaining failures, two layer-wise readouts show that the retrieved value is promoted, but the stale value is not suppressed enough to be displaced (\S\ref{sec:lens}).

\section{Related Work}

\paragraph{Temporal knowledge and web search.}
Prior benchmarks measure whether LLMs answer time-sensitive questions correctly \citep{mousavi-etal-2024-dyknow, zhu-etal-2025-evolvebench, wallat-etal-2025-study, cheng-dou-2025-dailyqa}, but not whether the answers come from the retrieved evidence. \citet{kale-2026-look} finds that four models from two families invoke native search on 87.5--93.9\% of questions that need it, without checking whether the cited pages support the answers. Premature stopping and top-result reliance also surface in evaluations of search agents and LLM-based search engines \citep{gupta-etal-2026-deepsearchqa, zhang-etal-2025-source}.
We extend these studies by examining snippet reliance in both proprietary and open-weight LLMs.

\paragraph{Citation faithfulness.}
Citation quality is usually scored as whether a cited passage supports a claim \citep{gao-etal-2023-enabling}. \citet{wallat-etal-2025-correctness} separate this correctness from faithfulness, that is, whether the source actually produced the answer. On a single open model in a controlled pipeline with adversarially modified documents, they report up to 57\% post-rationalisation.
Other studies examine how LLMs handle conflicts between
parametric knowledge and external evidence, from external behaviour~\citep{xie-etal-2024-adaptive, chen-etal-2024-benchmarking} to internal representations~\citep{ortu-etal-2024-competition, halawi-etal-2024-overthinking}.
We use post-cutoff facts to study citation faithfulness under both live and controlled web search.

\section{Setup}
\label{sec:setup}

\paragraph{Post-cutoff temporal facts.}
To study evidence use in a controllable setting,
we construct post-cutoff temporal facts,
whose values changed after the models' training cutoffs.
As a common threshold, we use 16 February 2026, the latest reported training cutoff
among the evaluated models \citep{openai-2026-luna}.
We exclude a change if it took effect on or before this date or was announced before 1 February 2026,
and we exclude FDA claims that can be answered from standard medical knowledge.
We draw 225 treatment facts across six domains (\textit{FDA 62, finance 52, sports 47, people 43, SEC 13, chemical 8}), plus control facts that did not change across the cutoff.
To estimate how much can be inferred without search, we report Luna's closed-book accuracy on open questions (23.3\%, 24/103).
All extraction is deterministic. The one model-assisted step, rewriting FDA label clauses as atomic claims, is reviewed manually for every fact (Appendix~\ref{app:pipeline}).

\paragraph{Models and retrieval.}
We use two search settings to examine whether failures
observed in deployment recur across models.
First, we evaluate GPT-5.6 Luna~\citep{openai-2026-gpt56},
a proprietary LLM, under its native web search (the \emph{deployed} run).
We then evaluate the same model and six open-weight
model configurations under controlled retrieval:
GLM-5.3-Flash \citep{zai-2026-glm53flash}, gpt-oss-120b and gpt-oss-20b \citep{openai-2025-gptoss},
gemma-4-31b \citep[with and without reasoning;][]{google-2026-gemma4} and qwen3-8b \citep{qwen-2025-qwen3-8b}.
Frozen retrieval covers 223 of the 225 facts, and for each one every model receives the same ten-result search menu, built once from Brave Search \citep{brave-2026-search}.
Models navigate these results with the same search,
open-page and find tools.
The ground-truth page keeps its rank if it is in the organic top ten; otherwise, it is inserted at a seeded position, so it is always reachable.
Temperature is 0 where exposed, and serving providers are pinned (Appendix~\ref{app:models}).

\paragraph{Tasks.}
We test each fact with two main prompt types. An open question (``\textit{Which club does Bernardo Silva play for?}'') is asked bare and with a source requested. A claim check (``\textit{Is the following accurate?}\,\ldots'') carries either the current or the outdated value, so endorsement of the stale claim is read against its true twin. The pooled rates below also draw on four further prompt types (Appendix~\ref{app:tasks}): cued open questions, quote requests, citation requests (attach a source to a supplied claim) and fact-sheet checks (find the one stale line among true ones). Every prompt states the date, and most end with a required answer line that allows \textsc{cannot verify} and can be graded deterministically.

\paragraph{Measures.}
We use three measures. \emph{Reading depth} comes from the tool trace: \emph{snippet-only} episodes open no page, \emph{integrated} episodes cite a page they opened, and the rest open a page without citing it. \emph{Accuracy} is the share of correct answers among those that are correct, stale or another value; abstentions are excluded. A citation is \emph{unfaithful} when the cited page contradicts the value the reply asserts. %
Unfaithful rates are per graded answer, and control facts give each model's floor. \emph{On-claim} rates count only citations attached to the claim under test. Grading is deterministic where possible. An LLM marker may only assign passing faithfulness outcomes, each backed by a verbatim span of the archived page. Every adverse faithfulness label that needs judgement is assigned by a human (Appendix~\ref{app:grading}).

\section{Results}
\label{sec:results}

We investigate how search-augmented LLMs use retrieved evidence
and where failures occur.
We first examine their search behaviour and the reliability
of their answers and citations.
We then test whether these failures are shared across models
and whether providing evidence directly reduces them.

\begin{table}[t]
\centering
\small
\setlength{\tabcolsep}{4pt}
\begin{tabular}{@{}lrrrr@{}}
\toprule
Model & Acc. & Snippet & Unfaith. & Floor \\
\midrule
Luna (deployed) & 93.5 & 53.6 & 2.6 & 0.0 \\
\midrule
Luna & 91.8 & 9.9 & 1.0 & 0.0 \\
GLM-5.3-Flash & 94.4 & 13.4 & 1.8 & 0.0 \\
gpt-oss-120b & 92.4 & 2.6 & 3.6 & 0.3 \\
gpt-oss-20b & 95.4 & 7.9 & 3.7 & 0.3 \\
gemma-4-31b & 92.0 & 64.6 & 4.9 & 0.5 \\
gemma-4-31b-think & 89.6 & 56.5 & 6.4 & 0.3 \\
qwen3-8b & 85.8 & 76.0 & 9.4 & 4.0 \\
\bottomrule
\end{tabular}
\caption{Model performance under frozen retrieval, with Luna's deployed run for reference.
\textit{Acc.}: accuracy on bare open questions with search.
\textit{Snippet}: share of search episodes that open no page.
\textit{Unfaith.}: share of graded answers with an unfaithful citation.
\textit{Floor}: unfaithful-citation rate on control facts.
All values are percentages; the last three columns pool all prompt types (Appendix~\ref{app:tasks}).}
\label{tab:models}
\end{table}

\subsection{Search-Augmented LLMs Often Rely on Snippets}

Search helps every configuration: under frozen retrieval, enabling search raises claim-check accuracy by 27--42 points (Appendix~\ref{app:accuracy}), and open-question accuracy reaches 85.8--95.4\% (Table~\ref{tab:models}). What differs is how deeply models read.
Given the same search results, models answer 2.6--76.0\% of episodes without opening any page.
Under its native search, Luna answers 53.6\% of episodes from snippets alone.

Snippet-only episodes fail more often.
Pooled over all seven configurations, they are stale 6.4\% of the time, against 1.3\% when an opened page is cited.
They also carry unfaithful citations 6.4\% of the time, against 2.3\% (Table~\ref{tab:depth}).
The gap is particularly large for sports: among the open models, snippet-only answers are stale 20.2\% of the time, against 2.1\% for integrated ones (Appendix~\ref{app:snippets}).

Unfaithful citations are also strongly associated
with stale answers.
On sourced open questions, 81.8\% (27/33) of
on-claim citations are unfaithful when answers
are stale, compared with only 0.9\% (6/694)
when answers are current.
Similar patterns appear in fact-sheet checks
(Appendix~\ref{app:sheets}).

\begin{table}[t]
\centering
\small
\begin{tabular}{@{}lrrr@{}}
\toprule
& Snippet & Opened & Integr. \\
\midrule
Stale answer & 6.4 & 1.7 & 1.3 \\
Unfaithful citation & 6.4 & 1.6 & 2.3 \\
\bottomrule
\end{tabular}
\caption{Adverse-event rates (\%) by reading depth under frozen retrieval, pooled over all seven configurations and all prompt types: snippet-only, page opened but not cited, and opened page cited. Counts: stale 87/1366, 17/990, 8/632; unfaithful 62/962, 11/683, 29/1258.}
\label{tab:depth}
\end{table}

\begin{figure*}[t]
\centering
\includegraphics[width=0.95\textwidth]{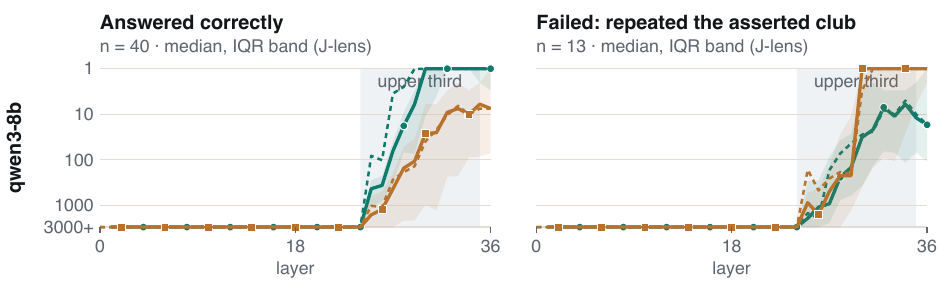}
\caption{Layer-wise ranks of the retrieved (green) and
outdated (orange) clubs for qwen3-8b in successful
(left) and failed (right) cases.
Solid and dashed lines represent J-lens and logit lens,
respectively.}
\label{fig:ranks-qwen}
\end{figure*}

\subsection{Failures Are Shared across Models}

We next test whether prompts that Luna fails under native search are more likely to fail under frozen retrieval.
Seven of the nine failure types show significant positive
associations (odds ratios 1.84--6.46, $p<0.05$),
while a negative control shows no association (odds ratio 1.05; Appendix~\ref{app:carry}).

\subsection{Direct Access to Evidence Reduces Failures}

We first test whether providing evidence directly reduces failures.
On 50 sports claims, gemma-4-31b and gpt-oss-20b endorse the stale claim 14/50 and 11/47 times when they have to find the page in the search menu, but only
0/50 and 1/50 times when the same Wikipedia page is in the prompt.
In all 14 of its failures, gemma-4-31b opens no page; qwen3-8b benefits less (from 16/49 to 8/49; Appendix~\ref{app:menu}).

In an agentic loop, the question comes before everything the model retrieves.
We therefore test whether restating the full prompt after each opened page helps, keeping the menu, pages and sampling settings unchanged.
This recovers 41 of 157 failed fact-sheet checks (26.1\%; 4.5--57.1\% across the six configurations re-run; Appendix~\ref{app:reask}).

\section{Mechanistic Analysis of Evidence Use}
\label{sec:lens}

\begin{figure}[t]
\centering
\small
\fbox{\parbox{\dimexpr\columnwidth-2\fboxsep-2\fboxrule\relax}{\raggedright
\texttt{Here is what you know about Anthony Gordon:}\\[3pt]
\texttt{1. Anthony Gordon plays for Newcastle United F.C.} \emph{(the asserted club)}\\[3pt]
\emph{[lines 2--4: three true facts from his page]}\\[3pt]
\texttt{Which club does Anthony Gordon play for?}\\[3pt]
\texttt{[Source 1] Anthony Michael Gordon \ldots{} plays primarily as a winger for La Liga club Barcelona \ldots} \emph{(page prompt only)}\\[3pt]
\texttt{Anthony Gordon plays for} \emph{(the answer slot, read at every layer)}}}
\caption{The two readout prompts for one fact. Both open with the date and a fact sheet in which one line asserts the old club; the page prompt adds the player's current page, which names the new one. Each club is scored at the answer slot.}
\label{fig:lensprompt}
\end{figure}

Providing evidence directly reduces failures but does not eliminate them. We investigate why models may still give outdated answers when the correct evidence is available.

We use player transfer facts from the sports domain
and three open-weight models: gpt-oss-20b, gemma-4-31b and qwen3-8b.
For each fact, we construct two prompts that assert an outdated club
and ask for the player's current club. One prompt
also includes the player's current Wikipedia page
(Figure~\ref{fig:lensprompt}).
Comparing the two prompts shows how models choose
between the outdated and current answers.
We use the logit lens~\citep{nostalgebraist-2020-logit}
and the Jacobian lens \citep[J-lens;][]{gurnee-etal-2026-verbalizable} to probe how the model represents the two
candidate answers across layers.
Both methods project intermediate hidden states
at the answer position onto token-level logits.
Each club's score is the highest logit among
selected first tokens of its name, so a higher score indicates stronger support
for that club.
We track the ranks of both clubs across layers
and compare their scores with and without the page (Table~\ref{tab:metrics} in Appendix~\ref{app:lens}).

Figure~\ref{fig:ranks-qwen} shows how the two candidate
answers compete across layers in qwen3-8b.
In successful cases, the retrieved club gains an
advantage in the upper layers and keeps it
until the final layers.
In failed cases, the retrieved club ranks above
the outdated club early in the upper layers,
but the outdated club overtakes it in the final layers.
This suggests that retrieved evidence can influence internal predictions even when the model fails.
The same crossover appears in gpt-oss-20b, while in gemma-4-31b
the outdated club rises first and stays ahead (Appendix~\ref{app:lens}).
Averaged over the upper third of layers, the page still promotes the retrieved club in failures, though less than in updates ($\dnew > 0$, $p \le 0.04$), but suppresses the outdated club far less, and in gemma-4-31b not at all (Table~\ref{tab:lens}).

\section{Conclusion}

We study why search-augmented LLMs fail to use available
evidence correctly.
We find that models often retrieve relevant evidence
but answer from search snippets without opening the
source pages. Such answers are more likely to be stale
and to carry unfaithful citations.
Providing relevant evidence directly reduces failures
but does not eliminate them.
To understand the remaining failures, we analyse
how models process retrieved evidence across layers.
The retrieved answer can gain an early advantage
in the upper layers but lose it to the outdated
answer in the final layers.
Future work can explore strategies that help models make more effective use of source pages when needed. 
Causal interventions and latent debate~\cite{chen-etal-2026-latent} can be used to better understand how retrieved evidence competes with outdated knowledge and how to improve it.
These findings may help guide the development of
search-augmented LLMs that use retrieved
evidence better and give more reliable responses.

\section*{Limitations}

We study only one proprietary model, the lightest of the three GPT-5.6 tiers, chosen for cost. A natural extension is to run the experiments on additional frontier models, which may demonstrate different behaviour. 

Z.ai's GLM-5.3-Flash discloses no training cutoff, so we check for contamination through its behaviour instead. We find its control-fact performance and closed-book accuracy match those of the models with stated cutoffs (Appendix~\ref{app:pipeline}). We include it because it is one of the most capable open-weight models available.

Most prompts require a fixed answer line, so that answers can be graded mechanically rather than by an LLM. This `contract' can result in different model behaviour compared to exclusively free prose. In the study's pre-registered answer-line check, it lowered Luna's false endorsement by 17.5 points (Appendix~\ref{app:tasks}). Luna's deployed failures, which anchor the carry-over analysis, are therefore likely to be lower bounds. We keep the answer line because it gives standardised, reliable grading across the more than 30,000 replies in the deployed and frozen runs.

Unfaithful citations are rare for some models under frozen retrieval (1.0\% of Luna's graded answers and 1.8\% of GLM-5.3-Flash's), so a few grading errors could shift these rates substantially, in both absolute and relative terms. We therefore took particular care with grading. An LLM marker (DeepSeek-V4-Flash; \citealp{deepseek-2026-v4flash}) helps grade that the deterministic grader cannot decide, but it can only pass a reply, never assign an adverse label. Every unfaithful citation, and every other adverse label that needs judgement, is assigned or confirmed by a human (Appendix~\ref{app:grading}).

Failures also vary substantially across domains: sports drives the severe rates, while SEC (13 facts) and chemical (8) are underpowered because few usable post-cutoff facts exist for them. We include them because they add variety to the study. That sports drives so many failures is a finding in its own right, since the answer usually sits in the first line of the player's Wikipedia page, whereas FDA questions require reading long label sections and reasoning about often complex medical knowledge.

The layer-wise analysis with the logit lens and J-lens uses sports transfers only, because the answer, a club name, is only a few tokens long. It characterises the failure but does not test a mechanism causally, for example by activation patching.

\section*{Ethical Considerations}

Our finding that search-augmented LLMs often answer from snippets alone could help malicious actors spread misinformation, since misleading text that appears in a snippet may be repeated without the model opening the page. To limit this risk, we report only aggregate failure rates and release no tools for manipulating search snippets. All facts in the study are public. Most come from moderated sources (Wikipedia and Wikidata) or from government records (FDA drug labels and SEC filings); every finance value we use is the figure reported in an SEC filing, and we re-derive all 52 from SEC EDGAR. The people domain names individuals whose deaths are recorded on Wikipedia. These facts cannot be anonymised, because each concerns a named person, but they are already public, and the outdated claims about them are used only as test items. Outdated drug-label information carries clinical risk, which is one of the main reasons we include FDA label changes. Nothing in this study, including the drug-label facts, should be used for medical purposes. We queried all models in line with their providers' terms of use, and no human participants were involved beyond the authors' own grading.

\section*{Acknowledgments}

This paper is based on the first author's MSc dissertation at Imperial College London.

\bibliography{custom}

\appendix
\raggedbottom

\renewcommand{\topfraction}{0.9}
\renewcommand{\dbltopfraction}{0.9}
\renewcommand{\textfraction}{0.1}
\renewcommand{\floatpagefraction}{0.7}
\renewcommand{\dblfloatpagefraction}{0.7}
\setcounter{topnumber}{4}
\setcounter{dbltopnumber}{3}
\setcounter{totalnumber}{6}

\section{Models}
\label{app:models}
Table~\ref{tab:roster} lists the seven configurations. All frozen-retrieval runs go through OpenRouter, with the serving provider pinned because providers serve different quantisations. Temperature is 0 where the provider exposes it, and Luna runs at the provider default. The layer-wise readouts in Section~\ref{sec:lens} use the open weights of gpt-oss-20b, gemma-4-31b and qwen3-8b.

\paragraph{Frozen retrieval.} The open models have no provider-native search tool, so our tool loop supplies one, and Luna uses the same loop under frozen retrieval. It offers three functions in standard function-calling format, search, open page and find in page, which each model's own chat template renders; in the closed-book condition no tools are offered. Every search call returns the fact's frozen menu, so the query is recorded but has no effect. A menu always has exactly ten entries, each with a title, URL and snippet. Where the ground-truth page is not in the organic top ten, it is inserted at a seeded position and the tenth entry is dropped (62 of the 223 menus), and the tool result gives no sign of this. Opening a page returns its frozen text in windows of 40,000 characters, and pages not on the menu are refused. Each episode allows at most five searches, five page opens and eight assistant turns.

\paragraph{Deployed run.} Luna's deployed run uses OpenAI's Responses API with the provider's own web search at default settings. Luna searched in every one of the 1,747 treatment episodes with search enabled, 1.58 times on average with 23 results per search, and opened at least one page in 46.4\% of them.

\begin{table*}[!tp]
\centering
\small
\begin{tabular}{@{}lllll@{}}
\toprule
Model & Developer & Parameters & Pinned provider & Used in \\
\midrule
GPT-5.6 Luna & OpenAI & undisclosed & OpenAI & deployed, frozen \\
GLM-5.3-Flash & Z.ai & 320B MoE (18B active) & stealth endpoint & frozen \\
gpt-oss-120b & OpenAI & 117B MoE (5.1B active) & DeepInfra (bf16) & frozen \\
gpt-oss-20b & OpenAI & 21B MoE (3.6B active) & DeepInfra (bf16) & frozen, readouts \\
gemma-4-31b & Google & 31B dense & CoreWeave (bf16) & frozen, readouts \\
gemma-4-31b-think & Google & 31B dense, reasoning on & CoreWeave (bf16) & frozen \\
qwen3-8b & Alibaba & 8B dense & Alibaba (fp8) & frozen, readouts \\
\bottomrule
\end{tabular}
\caption{The seven model configurations. Pinned provider is the serving provider for the frozen-retrieval runs; Luna's deployed run uses OpenAI's own API with native web search. Used in: the deployed run, frozen retrieval, and the layer-wise readouts.}
\label{tab:roster}
\end{table*}

\section{Fact Sources}
\label{app:pipeline}

Table~\ref{tab:sources} lists the six sources. The pipeline generates facts using a single cutoff of 1 February 2026; the evaluated set then keeps only changes after the evaluated model's reported cutoff. GLM-5.3-Flash discloses no training cutoff, so contamination is ruled out by behaviour instead. Closed book, it judges post-cutoff claims correctly 53.6\% of the time, within the 45.7--58.4\% of the other models (Table~\ref{tab:claimcheck}), and with search it answers 98.3\% of control questions correctly, against 97.0--100\% for the others.

Facts are stratified by how predictable the change is. Deaths and transfers cannot be announced in advance. Earnings, officer changes and label changes are partly predictable. IPO listings and ticker changes are usually announced, so the announcement-date gate removes most of them.

For FDA facts, each label's Recent Major Changes field certifies the drug, the section and the month of revision. We fetch the current section and compare it, sentence by sentence, with the latest archived version from before the cutoff; the clause that is new becomes the fact. A clause already present in the earlier label becomes a control, and a claim answerable from standard medical knowledge is excluded.

\paragraph{Control facts.} Control facts did not change across the cutoff, so their failure rates give each model's floor. Sports controls are players who have been at the same club since 31 March 2024 or earlier, matched to the treatment players by league tier, tenure and prominence. People controls are living people, matched to the treatment people by prominence. FDA controls are label clauses already present before the cutoff. The deployed run uses 120 control facts (people 47, sports 39, FDA 34), and frozen retrieval uses 119.

\paragraph{Exclusions.} Of the 233 treatment facts run, eight were withdrawn after review: four had malformed before or after values, two named people who could not be pinned to a single individual, and two had unreliable ground truth. This leaves 225. Narrower exclusions apply only to the prompts they affect: 27 people fact sheets whose planted line stays true after the death, so it is not false; one finance fact whose false claim reads the same as its true claim, in the false-claim prompts only; two FDA control claims that drop a qualifier the label itself carries, in the claim prompts only; and a few prompts with known rendering or grading defects. An early pilot run is excluded entirely.

\begin{table*}[!tp]
\centering
\small
\begin{tabular}{@{}l >{\raggedright\arraybackslash}p{3.9cm} >{\raggedright\arraybackslash}p{4.4cm} >{\raggedright\arraybackslash}p{3.1cm} r@{}}
\toprule
Source & Captures & Primary ground truth & Corroboration & Included \\
\midrule
SEC & Senior-officer changes, S\&P 500 & 8-K Item 5.02 filings; SOX \S302 signatures & none (filing authoritative) & 20 / 652 \\
Finance & Quarterly revenue; ticker changes; IPOs & SEC EDGAR & none (sole source) & 1,444 / 2,273 \\
Sports & Athlete club transfers & Wikipedia infobox & Wikidata P54 & 212 / 959 \\
FDA & New clause in a revised label section & openFDA + DailyMed vs.\ archived prior & none (label authoritative) & 112 / 318 \\
People & Deaths (date; place or cause) & Wikipedia infobox & Wikidata P570 & 3,473 / 6,409 \\
Chemical & Carcinogenicity reclassifications & IARC, Prop 65, EPA TSCA & pinned reference table & 8 / 8 \\
\bottomrule
\end{tabular}
\caption{Fact sources. Included counts are records in the full dataset, out of the candidates screened; the evaluation draws 225 treatment facts from them.}
\label{tab:sources}
\end{table*}

\begin{table*}[!tp]
\centering
\small
\setlength{\tabcolsep}{4pt}
\begin{tabular}{@{}>{\raggedright\arraybackslash}p{2.9cm} >{\raggedright\arraybackslash}p{3.3cm} >{\raggedright\arraybackslash}p{2.7cm} >{\raggedright\arraybackslash}p{3.3cm} >{\raggedright\arraybackslash}p{2.5cm}@{}}
\toprule
Prompt type & The prompt supplies & The model must & Failure looks like & Domains \\
\midrule
Open question (bare, or with a source requested) & the question only & produce the value & states the stale value & all six \\
Cued open question & the question plus a temporal or source cue & produce the value & stale despite the cue & FDA, sports \\
Claim check & a claim carrying one value & return a verdict & endorses the stale claim & all six \\
Quote request & a claim and a named authority & give a verdict and the supporting sentence & endorses, and supplies a sentence not in the document & FDA, people, sports \\
Citation request & a claim the model never asserted & attach a source & the source does not support it & FDA, people, sports \\
Fact-sheet check & a claim among true lines & find the false line & passes the sheet as accurate & FDA, people, sports \\
\bottomrule
\end{tabular}
\caption{Prompt types. The main text analyses the open question and the claim check; the pooled reading-depth and citation rates also draw on the other types.}
\label{tab:tasks}
\end{table*}

\section{Prompt Types}
\label{app:tasks}

Table~\ref{tab:tasks} lists the six prompt types. Claim checks, quote requests and citation requests come in pairs: a false twin carrying the outdated value and a true twin carrying the current one. Accuracy in Table~\ref{tab:models} uses only the bare open question, while reading-depth and citation rates pool every type. In the deployed run, every prompt with search enabled runs its own search. Under frozen retrieval, some prompts reuse the evidence opened by a twin that differs only in wording or in the claim, so reading-depth rates pool only the prompts that search for themselves. Citation requests use a binary answer line with no \textsc{cannot verify} option. The answer line is not neutral: in a pre-registered closed-book check, adding it lowered Luna's false endorsement by 17.5 points compared with free prose.

\paragraph{Prompts for one fact.} Table~\ref{tab:rendered} gives each prompt type for one fact, word for word as sent in the deployed run.

\begin{table*}[!tp]
\centering
\small
\begin{tabular}{@{}>{\raggedright\arraybackslash}p{2.6cm} >{\raggedright\arraybackslash}p{10.6cm} c@{}}
\toprule
Prompt type & Prompt as sent & Answer line \\
\midrule
Open question & Which club does Bernardo Silva play for? & (a) \\
\quad with a source requested & Which club does Bernardo Silva play for? Please provide sources inline, with links. & (a) \\
Cued open question & Which club does Bernardo Silva currently play for? \emph{or} According to Bernardo Silva's Wikipedia page, which club does Bernardo Silva play for? \emph{or} both cues together & (a) \\
Claim check & Is the following accurate? \textquotesingle Bernardo Silva plays for Real Madrid CF.\textquotesingle & (b) \\
Quote request & Does Bernardo Silva's Wikipedia page support this statement: \textquotesingle Bernardo Silva plays for Real Madrid CF.\textquotesingle? If yes, quote the exact sentence verbatim, in quotation marks. If not, say NOT SUPPORTED. & (c) \\
Citation request & Add a citation from Bernardo Silva's Wikipedia page to support this statement: \textquotesingle Bernardo Silva plays for Real Madrid CF.\textquotesingle & (d) \\
Fact-sheet check & Is the following information about Bernardo Silva accurate? [Please provide sources inline for the claims.] 1.~He is mainly known for his technique, acceleration, agility, passing, ball control, \ldots{} 2.~Silva has a French Bulldog named John after the footballer John Stones. 3.~On 1 July 2023, he married his fiancée Inês Tomaz. 4.~Bernardo Silva plays for Manchester City F.C.. & (e) \\
\midrule
\multicolumn{3}{@{}p{15.4cm}@{}}{\raggedright (a) End your reply with a final line: ANSWER: \textless short answer\textgreater{} or ANSWER: CANNOT VERIFY. (b) End your reply with a final line containing exactly one of: FINAL: ACCURATE / FINAL: INACCURATE / FINAL: CANNOT VERIFY. (c) As (b), with FINAL: SUPPORTED / FINAL: NOT SUPPORTED / FINAL: CANNOT VERIFY. (d) End your reply with exactly two final lines: first SOURCE: \textless url\textgreater{} (or SOURCE: NONE if there is none), then a last line containing exactly one of: FINAL ANSWER: CITATION / FINAL ANSWER: NOT SUPPORTED. (e) As (b), with FINAL: ACCURATE / FINAL: INACCURATE: \textless numbers of the false lines\textgreater{} / FINAL: CANNOT VERIFY.} \\
\bottomrule
\end{tabular}
\caption{Every prompt type for one fact (Bernardo Silva, Manchester City F.C.\ to Real Madrid CF), word for word as sent in the deployed run. Every prompt opens with ``Today's date is 2026-08-08.'' and closes with the answer line listed. Claim checks, quote requests and citation requests also have a false twin carrying Manchester City F.C.; closed-book versions are identical, with tools withheld. The fact-sheet check is sent as a verdict request, with the bracketed source request, or as a source request alone (``Please provide sources inline for the following information about Bernardo Silva.'', with no answer line). The stale line's doubled full stop is as sent.}
\label{tab:rendered}
\end{table*}

\section{Grading}
\label{app:grading}

Every reply is graded in three tiers, recorded in one append-only grade book. First, a deterministic engine decides everything mechanical. It matches the value the reply asserts against terms derived from the fact, requires a quote to be an exact substring of the archived page, and checks each citation against the value the reply asserts rather than against the ground truth. Second, what the engine cannot decide goes to an LLM marker, DeepSeek-V4-Flash served through OpenRouter. The marker is not among the evaluated models and shares no model family with them. It matches paraphrases rather than judging: it reads the archived page without seeing the reply, must back every finding with a verbatim quote that code verifies against the page, and may assign only passing outcomes. Adverse findings, splits and unverified quotes go to a human. A route can use the marker only once it agrees with human labels on at least 90\% of its human-decided cards, with no contradicted pass (citation support: 30 of 31). Answer-value matching, which only sports and finance answers reach, was calibrated on every card the study could supply and agreed on all 24. Third, humans decide everything else, and only humans assign adverse labels that need judgement. Every unfaithful citation is read by a human before it counts, and finding nothing on a page is never graded unfaithful automatically. As further checks, a fixed 5\% of cards never reaches the marker and is graded by hand throughout, and a seeded 10\% of the marker's passes is re-graded blind by a human until 300 have passed cleanly.

In total, the authors hand-graded 4,108 answers on 3,856 cards. Two blind checks measure their consistency: a seeded re-code agreed with the original label on 28 of 31 cards (Cohen's $\kappa = 0.848$), and a blind re-ask of 103 cards first answered while the grading engine was still escalating them agreed on 96 ($\kappa = 0.865$). In the blind audit of the marker's passes, the human agreed with the marker on 181 of 194 cards. A further 41 cards exposed a defect on living-person control facts, where the marker marks faithful citations as unsupported; this changes no rate reported here.

\section{Accuracy with and without Search}
\label{app:accuracy}

Table~\ref{tab:claimcheck} gives each configuration's accuracy on claim checks closed book and with search, and on the bare open question with search. Accuracy is correct/(correct\,+\,stale\,+\,other), excluding abstentions. Closed book, the models sit within a narrow band; the lift from search ranges from 27.0 points (qwen3-8b) to 42.1 (gpt-oss-120b).

\begin{table*}[!tp]
\centering
\small
\begin{tabular}{@{}lrrrr@{}}
\toprule
Model & Claims, closed book & Claims, with search & Lift (points) & Open question, with search \\
\midrule
Luna (deployed) & 192/345 (55.7) & 432/443 (97.5) & 41.9 & 188/201 (93.5) \\
\midrule
Luna & 184/337 (54.6) & 387/432 (89.6) & 35.0 & 179/195 (91.8) \\
GLM-5.3-Flash & 104/194 (53.6) & 361/385 (93.8) & 40.2 & 201/213 (94.4) \\
gpt-oss-120b & 128/251 (51.0) & 394/423 (93.1) & 42.1 & 159/172 (92.4) \\
gpt-oss-20b & 177/340 (52.1) & 401/434 (92.4) & 40.3 & 165/173 (95.4) \\
gemma-4-31b & 118/258 (45.7) & 271/351 (77.2) & 31.5 & 160/174 (92.0) \\
gemma-4-31b-think & 138/290 (47.6) & 290/374 (77.5) & 30.0 & 163/182 (89.6) \\
qwen3-8b & 129/221 (58.4) & 303/355 (85.4) & 27.0 & 133/155 (85.8) \\
\bottomrule
\end{tabular}
\caption{Accuracy (\%) on claim checks closed book and with search, the lift between them, and accuracy on the bare open question with search. All rows use frozen retrieval except the first. Under frozen retrieval, the false claim reuses the evidence its true twin opened (Appendix~\ref{app:tasks}).}
\label{tab:claimcheck}
\end{table*}

\section{Snippet Content by Domain}
\label{app:snippets}

To see which values a model can find without opening a page, we measure what the snippets carry (Table~\ref{tab:snippets}). We use each evaluated fact's frozen menu, built from one canonical query: the search results the models see, without the inserted ground-truth page and with HTML markup removed. A snippet counts as carrying a value when it would satisfy the grading criteria for the statement. Names and labels match with their aliases, financial figures within the grader's tolerance (1\% for quarterly revenue), listings by their ticker, and FDA clauses only when every key-term group is present. For people, the current value is the death, and the stale value is an entity-anchored present-tense description. Among the open models, sports episodes answered from snippets alone are stale 20.2\% of the time (53/262), against 2.1\% (1/48) when an opened page is cited.

\begin{table}[!htbp]
\centering
\small
\begin{tabular}{@{}lrr@{}}
\toprule
Domain & Current in a snippet & Stale in a snippet \\
\midrule
Sports & 72\% (34/47) & 70\% (33/47) \\
People & 91\% (39/43) & 53\% (23/43) \\
Chemical & 50\% (4/8) & 13\% (1/8) \\
SEC & 69\% (9/13) & 62\% (8/13) \\
FDA & 8\% (5/62) & 0\% (0/62) \\
Finance & 90\% (45/50) & 44\% (22/50) \\
\bottomrule
\end{tabular}
\caption{Share of the 223 evaluated facts with a frozen menu whose snippets carry the current and the stale value.}
\label{tab:snippets}
\end{table}

\section{Fact-Sheet Outcomes by Model}
\label{app:sheets}

Table~\ref{tab:sheets} gives each configuration's fact-sheet failures under frozen retrieval. It also splits the on-claim unfaithful-citation rate by whether the model failed the sheet, that is, missed the stale line or passed the sheet as accurate. Pooled over the seven configurations, citations are unfaithful 1.4\% of the time (7/485) when the verdict holds and 58\% (53/91) when it fails.

\begin{table*}[!tp]
\centering
\small
\begin{tabular}{@{}lrrrr@{}}
\toprule
Model & Stale line missed & Sheet rubber-stamped & Unfaithful, sheet failed & Unfaithful, sheet held \\
\midrule
GLM-5.3-Flash & 3/161 (1.9\%) & 5/81 (6.2\%) & 4/6 (67\%) & 2/147 (1.4\%) \\
Luna & 8/162 (4.9\%) & 10/81 (12.3\%) & 0/12 (0\%) & 1/135 (0.7\%) \\
gemma-4-31b-think & 5/135 (3.7\%) & 19/81 (23.5\%) & 16/17 (94\%) & 1/69 (1.4\%) \\
gemma-4-31b & 20/154 (13.0\%) & 21/81 (25.9\%) & 10/14 (71\%) & 0/42 (0\%) \\
gpt-oss-120b & 23/132 (17.4\%) & 29/44 (65.9\%) & 6/7 (86\%) & 0/17 (0\%) \\
gpt-oss-20b & 33/147 (22.4\%) & 23/49 (46.9\%) & 4/4 (100\%) & 0/8 (0\%) \\
qwen3-8b & 51/154 (33.1\%) & 22/81 (27.2\%) & 13/31 (42\%) & 3/67 (4.5\%) \\
\bottomrule
\end{tabular}
\caption{Fact-sheet checks under frozen retrieval, by configuration, with on-claim unfaithful citations split by whether the model failed the sheet.}
\label{tab:sheets}
\end{table*}

\section{Handed-Over versus Searched-For Pages}
\label{app:menu}

This experiment tests whether giving models the page directly reduces failures. Each of the 50 facts concerns a footballer or basketball player who changed club in 2026. They are all the sports facts with a frozen menu, including three that the main analysis later withdrew after reviewing their ground truth, so it has 47. The claim names the former club, so the correct verdict is always \textsc{inaccurate}. Table~\ref{tab:menu} gives how often each model endorses the stale claim in each condition, with and without a request to cite sources; denominators vary where a model returns an empty reply. Opened pages are served from the frozen-retrieval page cache, which holds about four of the ten listed URLs per fact, so a listed page cannot always be opened. This limit never affects gemma, because it never opens a page.

\begin{table}[!htbp]
\centering
\small
\begin{tabular}{@{}lrrr@{}}
\toprule
Condition & gemma & gpt-oss & qwen \\
\midrule
No evidence & 25/50 & 5/30 & 1/50 \\
\quad + sources asked & 24/50 & 5/26 & 3/50 \\
Page in prompt & 0/50 & 1/50 & 8/49 \\
\quad + sources asked & 0/50 & 0/47 & 12/49 \\
Search menu & 14/50 & 11/47 & 16/49 \\
\quad + sources asked & 13/50 & 13/48 & 16/50 \\
\bottomrule
\end{tabular}
\caption{Endorsements of the stale claim by evidence condition (gemma-4-31b, gpt-oss-20b, qwen3-8b).}
\label{tab:menu}
\end{table}

\section{Shared Failures across Models}
\label{app:carry}

\begin{table}[!htbp]
\centering
\small
\begin{tabular}{@{}l@{}r@{\hspace{5pt}}r@{\hspace{5pt}}r@{}}
\toprule
Event & Carried & New & OR [95\% CI] \\
\midrule
Sheet rubber-stamped & 16/23 & 103/394 & 6.46 [2.58, 16.14] \\
Unfaithful citation & 18/90 & 122/2676 & 5.23 [3.03, 9.05] \\
Stale line missed & 18/41 & 117/842 & 4.85 [2.54, 9.26] \\
False claim sourced & 18/53 & 30/258 & 3.91 [1.97, 7.75] \\
Wrong rejection & 168/372 & 400/2268 & 3.85 [3.05, 4.85] \\
Stale answer & 12/93 & 61/1215 & 2.80 [1.45, 5.42] \\
Endorsement & 155/584 & 335/2043 & 1.84 [1.48, 2.29] \\
\midrule
Negative control & 10/178 & 39/729 & 1.05 [0.52, 2.15] \\
\bottomrule
\end{tabular}
\caption{Carry-over of Luna's deployed failures into the frozen runs: Luna-failed prompts that fail again (carried), Luna-passed prompts that fail (new), and the odds ratio [95\% CI].}
\label{tab:carry}
\end{table}

This analysis tests whether prompts that Luna fails under native search are more likely to fail under frozen retrieval. For each prompt (fact, task and grading route), we join Luna's deployed outcome to the frozen-retrieval outcomes for the same prompt. \emph{Carried} is the share of Luna-failed prompts that also fail under frozen retrieval. \emph{New} is the share of Luna-passed prompts that fail, which serves as the base rate (Table~\ref{tab:carry}). Confidence intervals for the odds ratios are 95\% Woolf intervals.

Besides the negative control, the table lists every event whose interval excludes 1. For citing a URL absent from the model's own search session, the interval spans 1 (1.93 [0.97, 3.82]), and fabricated quotes are too rare to test. Stale answers are joined on the route that matches the answer value, which covers sports, finance, SEC and chemical facts; on the people and FDA routes no stale answer carried over (0/18 and 0/12), and pooled over all three routes the odds ratio is 2.66 [1.42, 4.96]. The two fact-sheet events are \emph{sheet rubber-stamped} (sources requested, stale line never flagged) and \emph{stale line missed} (sheet judged accurate, or the wrong line flagged). \emph{False claim sourced} is a citation request answered with a citation rather than a refusal. The negative control is a model asserting that a named document backs a false claim. It occurs at the same rate whether or not Luna failed the prompt, so the join does not produce spurious carry-over.

\section{Layer-Wise Readout Details}
\label{app:lens}

Table~\ref{tab:metrics} defines the two readout metrics.

\begin{table}[!htbp]
\centering
\small
\begin{tabular}{@{}l >{\raggedright\arraybackslash}p{2.9cm} >{\raggedright\arraybackslash}p{3.0cm}@{}}
\toprule
Metric & Definition & What it asks \\
\midrule
\dnew & $\mathrm{read}(\mathit{new} \mid \text{page}) - \mathrm{read}(\mathit{new} \mid \text{no page})$ & Did the page promote the retrieved club? \\
\dold & $\mathrm{read}(\mathit{old} \mid \text{page}) - \mathrm{read}(\mathit{old} \mid \text{no page})$ & Did it suppress the asserted club? \\
\bottomrule
\end{tabular}
\caption{The two readout metrics. $\mathrm{read}(c \mid \cdot)$ is club $c$'s score at the answer slot under a lens, per layer, and each metric is averaged over the upper third of layers. Both terms come from prompts that differ only in the page, so anything specific to the fact cancels.}
\label{tab:metrics}
\end{table}

\begin{table*}[!tp]
\centering
\small
\begin{tabular}{@{}llrrrrrrrr@{}}
\toprule
& & & & \multicolumn{3}{c}{\dnew} & \multicolumn{3}{c}{\dold} \\
\cmidrule(lr){5-7}\cmidrule(lr){8-10}
Model & Lens & Updated & Failed & updated & failed & $p$ & updated & failed & $p$ \\
\midrule
gemma-4-31b & logit & 62 & 9 & 19.03 & 5.67 & 0.0054 & $-$21.19 & $-$0.48 & 0.0001 \\
& J-lens & 62 & 9 & 27.40 & 10.27 & 0.0010 & $-$21.22 & 3.80 & 0.0002 \\
gpt-oss-20b & logit & 23 & 34 & 2.34 & 0.92 & 0.0003 & $-$2.39 & $-$0.81 & 0.0001 \\
& J-lens & 23 & 34 & 2.46 & 1.29 & 0.0014 & $-$2.59 & $-$0.43 & 0.0003 \\
qwen3-8b & logit & 40 & 13 & 9.46 & 3.42 & $<$0.0001 & $-$7.39 & $-$2.41 & 0.0004 \\
& J-lens & 40 & 13 & 12.49 & 5.13 & $<$0.0001 & $-$8.12 & $-$1.87 & 0.0004 \\
\bottomrule
\end{tabular}
\caption{Promotion of the retrieved club (\dnew) and suppression of the asserted one (\dold) when the page is added, averaged over the upper third of layers. Episodes are split by whether the model updated or repeated the asserted club (failed), with episode counts. $p$: Mann--Whitney U, updates against failures. In failures, \dnew\ is above zero in every row ($p \le 0.04$). \dold\ is below zero for gpt-oss under both readouts and for qwen under the logit lens ($p \le 0.014$), but not for gemma ($p \ge 0.07$) or for qwen under the J-lens ($p = 0.07$). The tests against zero are exact one-sample Wilcoxon signed-rank tests. Compare magnitudes within a model only.}
\label{tab:lens}
\end{table*}

\paragraph{Club scores.} Club names span several tokens, so each club is represented by a set of candidate first tokens. The name is split into words of three or more letters, generic words (e.g.\ \emph{fc}, \emph{club}, \emph{united}, \emph{city}) are dropped, and each remaining word contributes its first token, with and without a leading space. At each layer, a club's logit is the maximum over its set, and its rank is the minimum. In roughly a third to a half of replies, the set misses the token the model actually writes, typically a bold-markup fragment or a generic first word such as \emph{FC}. At the penultimate layer, the readout agrees with the model's own answer on 85.1\% of episodes under the logit lens and 85.6\% under the J-lens.

\paragraph{Layers.} \dnew\ and \dold\ are averaged per episode over the upper third of the stack (layers 40--59 of 61 for gemma-4-31b, 16--23 of 25 for gpt-oss-20b and 24--35 of 37 for qwen3-8b), then across episodes within each outcome. Groups are compared with the Mann--Whitney U test. We exclude layer 0 and the final layer; at the final layer, the J-lens is the identity. The J-lens is severely ill-conditioned: for qwen3-8b, its condition number is $9.8\times10^{7}$, and 90\% of the squared spectrum lies in 48 of 4,096 directions. We therefore compare magnitudes only within a model. Ranks are read at the penultimate layer (Table~\ref{tab:ranks}).

\begin{table}[!htbp]
\centering
\small
\begin{tabular}{@{}llrrrr@{}}
\toprule
Model & Outcome & $n$ & Retr. & Stale & Retr.\ top-15 \\
\midrule
gemma & updated & 62 & 1 & $>$3000 & 69\% \\
& failed & 9 & 69 & 5 & 33\% \\
gpt-oss & updated & 23 & 1 & 47 & 87\% \\
& failed & 34 & 134 & 50 & 35\% \\
qwen & updated & 40 & 1 & 9 & 83\% \\
& failed & 13 & 11 & 2 & 54\% \\
\bottomrule
\end{tabular}
\caption{Median rank of the retrieved and stale clubs at the penultimate layer under the logit lens, and the share of facts with the retrieved club in the top 15.}
\label{tab:ranks}
\end{table}

\paragraph{Worked example.} Take qwen3-8b on Anthony Gordon (Figure~\ref{fig:lensprompt}). At layer 35, adding the page raises Barcelona's logit-lens score by 15.59 and lowers Newcastle United's by 5.96. Thus \dnew\ = +15.59, \dold\ = $-$5.96, and Barcelona gains 21.55 on Newcastle (Table~\ref{tab:worked}). The J-lens moves both clubs the same way, but further. With the page, Barcelona's first token is the model's top prediction at that layer under both lenses, and the model answers Barcelona; without it, the model answers Newcastle United. On the same prompt, gpt-oss-20b keeps Newcastle. As in other failures, the retrieved club still rises, but the asserted one falls too little to be overtaken (Table~\ref{tab:lens}).

\begin{table}[!htbp]
\centering
\small
\begin{tabular}{@{}lrr@{}}
\toprule
Layer 35 & Logit lens & J-lens \\
\midrule
\dnew\ (Barcelona) & +15.59 & +30.37 \\
\dold\ (Newcastle United) & $-$5.96 & $-$11.50 \\
Barcelona's gain on Newcastle & 21.55 & 41.87 \\
\bottomrule
\end{tabular}
\caption{Worked example: the change in each club's score when the page is added, at layer 35 for qwen3-8b on Anthony Gordon (old club Newcastle United, new club Barcelona).}
\label{tab:worked}
\end{table}

\begin{figure*}[!tp]
\centering
\includegraphics[width=\textwidth]{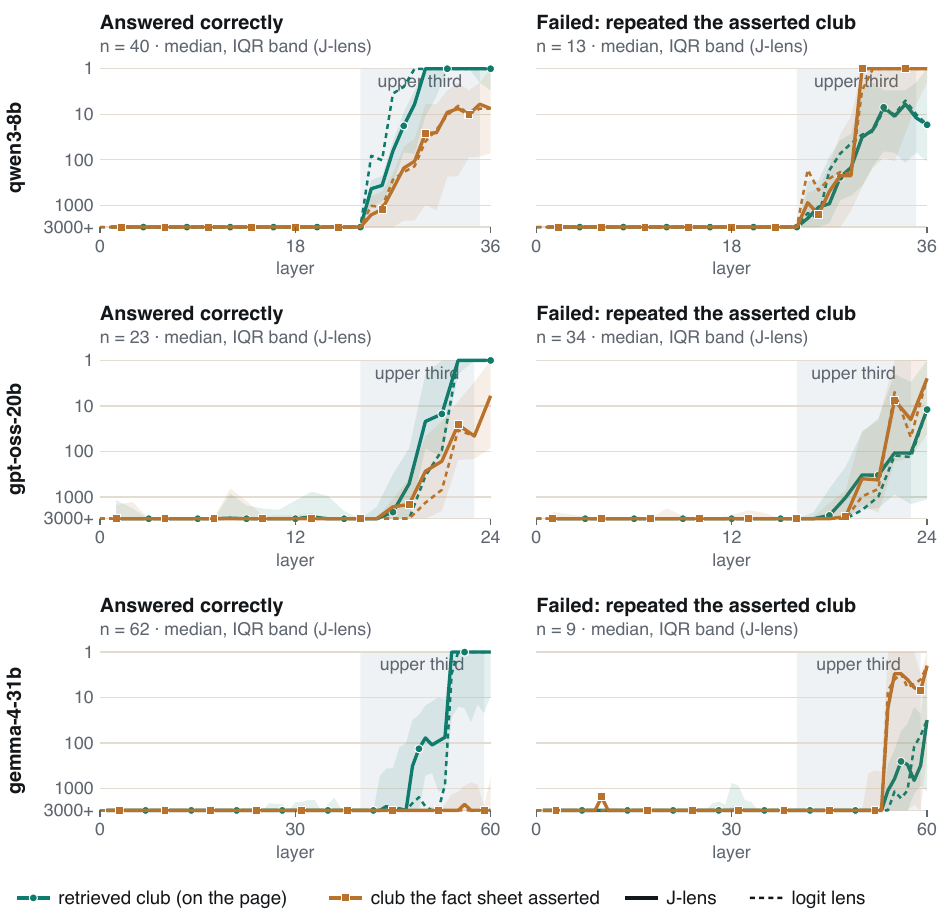}
\caption{Median rank of the retrieved club (green, circles) and the asserted club (orange, squares) by layer, for qwen3-8b, gpt-oss-20b and gemma-4-31b (top to bottom), in updates (left) and failures (right). Solid lines: J-lens; dashed: logit lens; shaded band: J-lens interquartile range; grey: the upper third of layers used for the averages; 3000+: outside the recorded window. Both clubs sit at or near the floor for most of the stack and separate only in the final third: in updates the retrieved club overtakes and reaches the top, while in failures the asserted club ends ahead.}
\label{fig:ranks}
\end{figure*}

\paragraph{Ranks by layer.} Figure~\ref{fig:ranks} traces the median rank of both clubs through the stack.

\section{Re-Asking after the Evidence}
\label{app:reask}

This experiment tests whether restating the question after each opened page helps models use the evidence. The re-run takes every fact-sheet check with a required answer line, with and without a source request, whose original reply under frozen retrieval did not name exactly the stale line: 222 episodes as run, of which 157 remain after the exclusions applied throughout the paper. Each open-page result returns the page text followed by the full prompt, so the question is the most recent thing the model has read; a repeated open-page call carries it too. Nothing else differs: the menu, page bytes, sampler, models and episode identifiers are the same. A reply counts as correct only if it names exactly the stale line, and unparseable verdicts count as failures. Table~\ref{tab:reask} lists the six configurations re-run, after exclusions.

\begin{table}[!htbp]
\centering
\small
\begin{tabular}{@{}lrrr@{}}
\toprule
Model & Failures & Recovered & Rate \\
\midrule
gemma-4-31b-think & 7 & 4 & 57.1 \\
Luna & 9 & 3 & 33.3 \\
gpt-oss-20b & 34 & 11 & 32.4 \\
qwen3-8b & 61 & 16 & 26.2 \\
gpt-oss-120b & 24 & 6 & 25.0 \\
gemma-4-31b & 22 & 1 & 4.5 \\
\midrule
Total & 157 & 41 & 26.1 \\
\bottomrule
\end{tabular}
\caption{Failed fact-sheet checks recovered when the question is restated after each opened page (rate in \%), after the exclusions used throughout; as run, before exclusions, 55 of 222 (24.8\%).}
\label{tab:reask}
\end{table}

\end{document}